\documentclass[eat,twocolumn]{jmlr}
\usepackage{longtable}
\usepackage[font=small,skip=-2pt,belowskip=-10pt]{caption}
\pdfoutput=1
\jmlrvolume{}
\firstpageno{1}
\jmlryear{2022}
\jmlrworkshop{Machine Learning for Health (ML4H) 2022}

\title[H-BERT]{Semantic Decomposition Improves Learning of Large Language Models on EHR Data\titletag{}}

\author{\Name{David A Bloore} \Email{dbloore@eqrx.com}\\
   \Name{Romane Gauriau} \Email{rgauriau@eqrx.com}\\
   \Name{Anna L Decker} \Email{adecker@eqrx.com}\\
   \Name{Jacob Oppenheim} \Email{joppenheim@eqrx.com}\\
\addr EQRx Inc.  50 Hampshire St. Cambridge MA 02139}

\begin{document}

\maketitle

\begin{abstract}
Electronic health records (EHR) are widely believed to hold a profusion of actionable insights, encrypted in an irregular, semi-structured format, amidst a loud noise background.  To simplify learning patterns of health and disease, medical codes in EHR can be decomposed into semantic units connected by hierarchical graphs.  Building on earlier synergy between Bidirectional Encoder Representations from Transformers (BERT)  and Graph Attention Networks (GAT), we present H-BERT, which ingests complete graph tree expansions of hierarchical medical codes as opposed to only ingesting the leaves and pushes patient-level labels down to each visit.  This methodology significantly improves prediction of patient membership in over 500 medical diagnosis classes as measured by aggregated AUC and APS, and creates distinct representations of patients in closely related but clinically distinct phenotypes.
\end{abstract}
\begin{keywords}
EHR, BERT, GAT
\end{keywords}

\section{Introduction}
\label{sec:intro}

EHR have a structure all their own, and taking advantage of that structure has proven exceedingly difficult.  Early approaches based on SkipGram \citep{mikolov2013efficient}, and even recent approaches based on BERT \citep{BERT} or similar architectures are not used in the clinic---whereas computer vision is becoming essential to the standard of care in clinical oncology.  EHR datasets are causally and temporally confounded, which makes treatment and outcome prediction and evaluation difficult and hard to interpret.  Disease phenotyping is a useful and meaningful task that can be considered upstream of such prediction tasks and provides insights regarding patient journeys through the healthcare system as well as disease etiology \citep{deepPhe, mixEHR}.

Phenotyping is \emph{de facto} upstream of any prediction, because an ML model must have different internal representations of patients associated with different outcomes.  The limits of phenotyping inform the limits of prediction: demonstrating where a model differentiates between patients can validate or help diagnose error in prediction results.

Our core focus is learning patient representations that allow us to make quantitative comparisons between patients in the patient embedding space.  Semi-supervised phenotyping and segmenting of patients using embeddings derived from EHR can be a starting place for characterization similarities and differences between patient groups to broadly support healthcare.

Medical codes themselves have a structure; they are always subclasses of more general concepts.  For example, the International Classification of Diseases (ICD) ICD-10 code I21.02 is for ``Acute Myocardial Infarction involving left anterior descending coronary artery,'' and is also:  a disease of the circulatory system, a type of ischemic heart disease, and so on.  {Shared hierarchical classes define informative relationships between codes.}

H-BERT is an approach to {both data and model}, decomposing medical codes into hierarchical graphs---and feeding those entire graphs into a neural network architecture designed specifically to exploit graphs of semantic tokens.  The full semantic decomposition of any concept is more informative than just the leaf token, because membership in hierarchical classes adds context and the meaning of any symbol is only defined by context.  We believe this combined approach simplifies, accelerates, and deepens learning.

\figureref{fig:sem_decomp} depicts semantic decomposition of I20.02 graphically.

\begin{figure}[htbp]
 \floatconts
  {fig:sem_decomp}
  {\caption{Semantic decomposition of ICD-10 code ``I21.02.''}}
  {\includegraphics[width=1\linewidth]{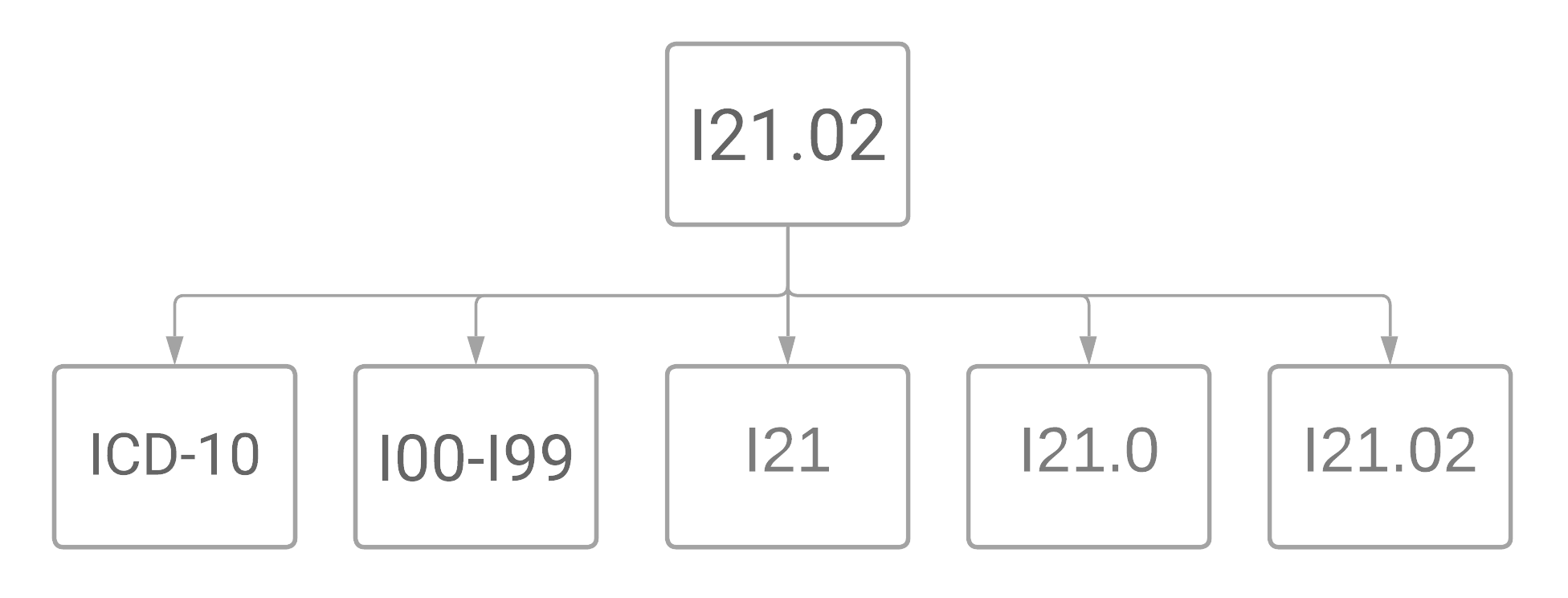}}
\end{figure}

\section{Background}
Given BERT's success on many different applications in NLP, many authors have constructed variations of BERT for use on EHR.  Four of the most relevant of these are:  G-BERT \citep{ijcai2019p825}, BEHRT \citep{Li:2020wz}, Med-BERT \citep{Rasmy:2021ug}, and CEHR-BERT \citep{CEHRBERT}. 

BEHRT, Med-BERT, and CEHR-BERT focus on different adaptations of a BERT-only model.  BEHRT and Med-BERT used diagnoses only and larger datasets than G-BERT.  CEHR-BERT focused on adding temporal tokens, and ingests diagnosis, prescription, and procedure codes.  The creators of CEHR-BERT also present a comprehensive ablation study comparing itself to BEHRT, Med-BERT, and others finding that none was best for all of the small set of very specific tasks.

G-BERT and an earlier work called GRAM \citep{gram} use variations of graph attention.  GRAM has a bespoke graph attention implementation feeding a GRU.  Notably, GRAM uses medical code initializations based on SkipGram, which improve their results beyond the random initializations they tried.  G-BERT uses an original GAT \citep{velickovic2018graph} feeding a BERT, and is the only example we know of in literature to do so.  ICD-9 diagnosis and Anatomical Therapeutic Chemical (ATC) ATC-4 prescription codes are processed by two separate GAT/BERT stacks, and combined downstream.  ICD-9 and ATC-4 graph trees are used by the GAT to execute message passing between tokens at adjacent levels of hierarchy.  G-BERT operates at the visit-level, and aggregates embeddings for medication recommendation---where it outperformed older methods (GRAM and RETAIN \citep{choi2017retain}, for example) by a narrow margin.  While novel and inspiring, their result is also limited by its reliance upon the small and unrepresentatively fine-grained MIMIC-III dataset \citep{MIMIC3}.  We believe that training on large EHR datasets from hospital systems and, similarly, closed claims datasets from entire payors present a much more real world relevant task for ML modelling.

The tasks in relevant literature are either prediction of future medical codes or phenotype membership.  GRAM focused on sequential diagnosis prediction and patient-level heart failure prediction, which is a set of tasks similar in nature to those of this work because there is a broad task and a specific task.  The tasks of the BERT-based models tended to be either very general or fairly specific---but not both.  G-BERT aimed to predict what will be prescribed given a history and current visit's diagnoses.  BEHRT sought to predict diagnosis codes in the next visit, within the next six months, and within the next 12 months.  Med-BERT targeted phenotyping, but only used two phenotypes and emphasized the clinical description of the phenotypes in their narrative.  CEHR-BERT used four phenotypes, mostly drawing from several prior works.

In any case, all of the phenotypes of prior work are clinically relevant:  heart failure, onset of cancer, heart failure in diabetics, etc.  The next medical code or prescription tasks are less relevant because they aren't very accurate, and might never be.  The next diagnosis or prescription might not be deterministically bound to prior diagnoses and prescriptions, e.g.\ acute injury or infection in a previously healthy subject.  Medication prediction is particularly troublesome because a model could learn outdated practice and will always underperform the clinical reality that is its training data.  We hypothesize, however, that these tasks are a valuable means to improve learning phenotypes, offering both a different task from pre-training and many more opportunities for learning than rarefied labels associated with narrowly defined phenotypes.

Aside from training tasks, we believe that there is room for improvement in leveraging the hierarchical nature of medical codes and improvements in large language models.   None of these models ingest semantic decomposition of medical codes, information that should be directly relevant for Large Language Model (LLM)-based architectures.  By using phenotyping tasks as a basis for comparison we hope to understand both the effects of sematic decomposition and architectural differences between models.  Ultimately, our performance on phenotyping bounds our ability to use EHR for predictive modeling and improved performance characterization will enable us to make progress towards that goal.

\section{Methods}

\subsection{Data}

The EHR data used in this work comes from Geisinger Health, a regional healthcare provider and payer in Pennsylvania.  Our final dataset includes 1,756,380 patients, and 30,353,712 visits (see visit definition below), further detail in \appendixref{apd:raw_data}.  The clinical experience captured is very broad, including:  healthy patients, outpatient visits, short and long-term inpatient stays, and treatment of rare diseases.

Raw data processing began with mapping diagnostic codes to ICD-10 and prescription codes to ATC-4.  Both of these code systems have hierarchies well-suited for this work.  After code mapping, patients' records are grouped into ``visits,'' defined as everything that happened on a given calendar day.  The visits used were further down-selected to only those containing at least one diagnosis code and at least one prescription code (mostly to keep the dataset tractable with available resources).  The training, validation, test split ratio is 70:10:20.
Semantic decomposition of each medical code was performed after down-selection to the final set of visits.  
Each medical code was also truncated according to a global maximum hierarchy depth for that code system \citep{codeTrunc}, as described in \appendixref{apd:codeTrunc}. 

Patient-level labels are constructed by two approaches:  (1) using clinician-designed cohort definitions as described in \appendixref{apd:cohorts}, or (2) by asking, ``did this patient ever have a diagnosis in this Clinical Classifications Software Refined (CCSR) category?'' for each of the 519 CCSR categories.  Once the patient-level labels are created, each visit of a patient gets that patient's labels.  The large number of relatively \emph{simple but related} tasks is intended to soften the learning curve for the model; multitask learning improves generalization \citep{multi}, while spanning CCSR makes all patients and diagnosis codes relevant to at least some predictions.  The specific and narrow cohort tasks provide a special challenge for the model, since the rules for cohort inclusion are more complex and membership is so low.  Of 1,756,380 patients, there are only 6,082 Atopic Dermatitis (AD), 1,144 Ankylosing Spondylitis (AxSpA), and 10,607 Rheumatoid Arthritis (RA) cohort patients. 

\subsection{Architecture}
H-BERT ingests the union of the semantic decomposition of all medical codes in a visit.  All inputs feed one GAT/BERT model---whereas G-BERT partitions diagnosis and prescription codes into separate GAT/BERT models.  H-BERT uses a newer GAT with an improved attention mechanism \citep{GATv2}.  \figureref{fig:eqb_arch} illustrates the difference between feeding only leaf tokens to a BERT-type model and the overall flow of information in H-BERT.

\begin{figure}[htbp]
\floatconts
  {fig:eqb_arch}
  {\caption{The information presented to the model is significantly different when (a) presenting only the leaf-token, compared to (b) presenting semantic decompositions of medical codes (or any other decomposable symbol.}}
  {%
    \subfigure[Leaf token ingestion only]{\label{fig:berto}%
      \includegraphics[width=0.5\linewidth]{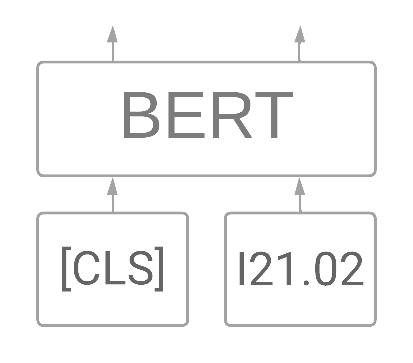}}%
    \qquad
    \subfigure[Semantic decomposition ingestion by H-BERT]{\label{fig:eqb}%
      \includegraphics[width=0.9\linewidth]{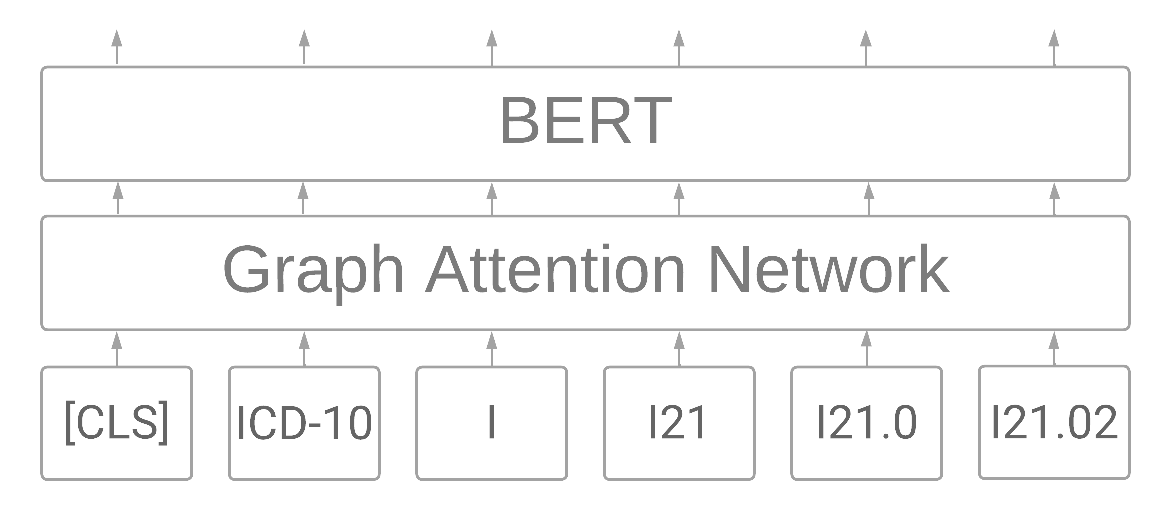}} 
  }
\end{figure}

Pre-training is via Masked Language Modeling (MLM)---as with BERT.  15\% of inputs are subject to loss calculation.  Of those, 80\% are replaced with a \verb'[MASK]' token, 10\% are replaced with a random token, and the remainder are left as is.

The fine-tuning task as implemented asks, ``is this visit from a patient who has a given disease label?''  By pushing patient-level labels down to the visit-level, this fine-tuning task achieves a simple form of longitudinality.

Patient-level embeddings are the average of all visit-level embeddings for that patient.  Here, we again drew inspiration from BERT, which sums its input segment, position, and token embeddings.  Since the number of visits per patient varies, it made more sense to use an average of visit embeddings than a sum.  The simplicity of this approach makes it especially practical.  Using an RNN, GRU, or other aggregation method as suggested in the literature may add value, but it also adds complexity.  The modular definition of these embeddings allows for natural extensions to phenotype prediction and explanation tasks in the future.

\section{Results}
Four configurations have been evaluated: (1) H-BERT ingesting semantically decomposed tokens by a GAT/BERT model (HB), (2) BERT-only ingesting semantically decomposed tokens (BERTO), (3) BERT-only ingesting leaf tokens only (LEAFO), and (4) LEAFO6 which is the same as LEAFO but uses 6 Transformer blocks versus LEAFO's 4 to have a number of trainable parameters between HB and BERTO.  We designed these comparisons to isolate the effect of (1) semantic decomposition, (2) graph attention, and (3) number of parameters to isolate the effects of architectural changes.  Of these, HB has the largest number of trainable parameters, and is the only configuration that has non-trainable parameters (which define the edge connections between nodes in the ICD-10 and ATC-4 graph trees).  Hyperparameters for reported results are discussed in \appendixref{apd:hyper}.  Duration of training was held constant for all models.

For each model, the AUC and APS was calculated for each task using Scikit-Learn \citep{scikit-learn}, after which we computed the simple arithmetic mean of the per-task AUC and APS scores. \tableref{tab:auc_aps} shows HB achieved the best mean AUC and APS of these four.  BERTO was second, but closer in performance to LEAFO than HB.  BERTO outperforming LEAFO suggests that semantic decomposition by itself improves learning.  And since HB best associates visits with labels, the combination of semantic decomposition and GAT outperforms just semantic decomposition absent a GAT.  This suggests that GATs are better designed to interpret semantic decompositions of symbols than Transformer blocks.  LEAFO6, with more trainable parameters, performed worst, suggesting the improved performance of HB and BERTO is due to graph attention architecture rather than model size.

\begin{table}[htbp]
\floatconts
  {tab:auc_aps}
  {\caption{Mean AUC and APS for H-BERT, BERTO, and LEAFO.}}%
  {%
    \begin{tabular}{|c|c|c|}
    \hline
    \abovestrut{2.2ex}\bfseries Cfg. & \bfseries $\mu_{AUC}$ & \bfseries $\mu_{APS}$ \\\hline
    \abovestrut{2.2ex} \textbf{HB} & \textbf{0.9099} & \textbf{0.2343} \\
    BERTO & 0.8894 & 0.2091 \\
    LEAFO & 0.8889 & 0.2077 \\
    \belowstrut{0.2ex}LEAFO6 & 0.8822 & 0.1945 \\\hline
    \end{tabular}
  }
\end{table}

Learning and distinguishing between specific, clinician-defined cohorts of patients enables many applications in clinical development, clinical operations, outcome prediction, and more.  \figureref{fig:pte_eqb} shows a PCA projection of patient-level embeddings produced using H-BERT.  The overall dataset is blue, with the AD and RA cohorts in orange and red, respectively.  The orange and red dots have clearly different distributions, indicating that H-BERT learns and expresses measurably different representations for these two related, similarly treated, but clinically distinct diseases.  \figureref{fig:pte_berto} shows BERTO does not produce patient-level embeddings for AD and RA patients that are demonstrably different, and LEAFO (see \appendixref{apd:leafo_pte}) does not either.

\begin{figure}[htbp]
\floatconts
  {fig:pte_eqb}
  {\caption{PCA projection of top two components of HB Patient Embeddings.  Note clear separation of AD and RA cohorts.}}
  {\includegraphics[width=1\linewidth]{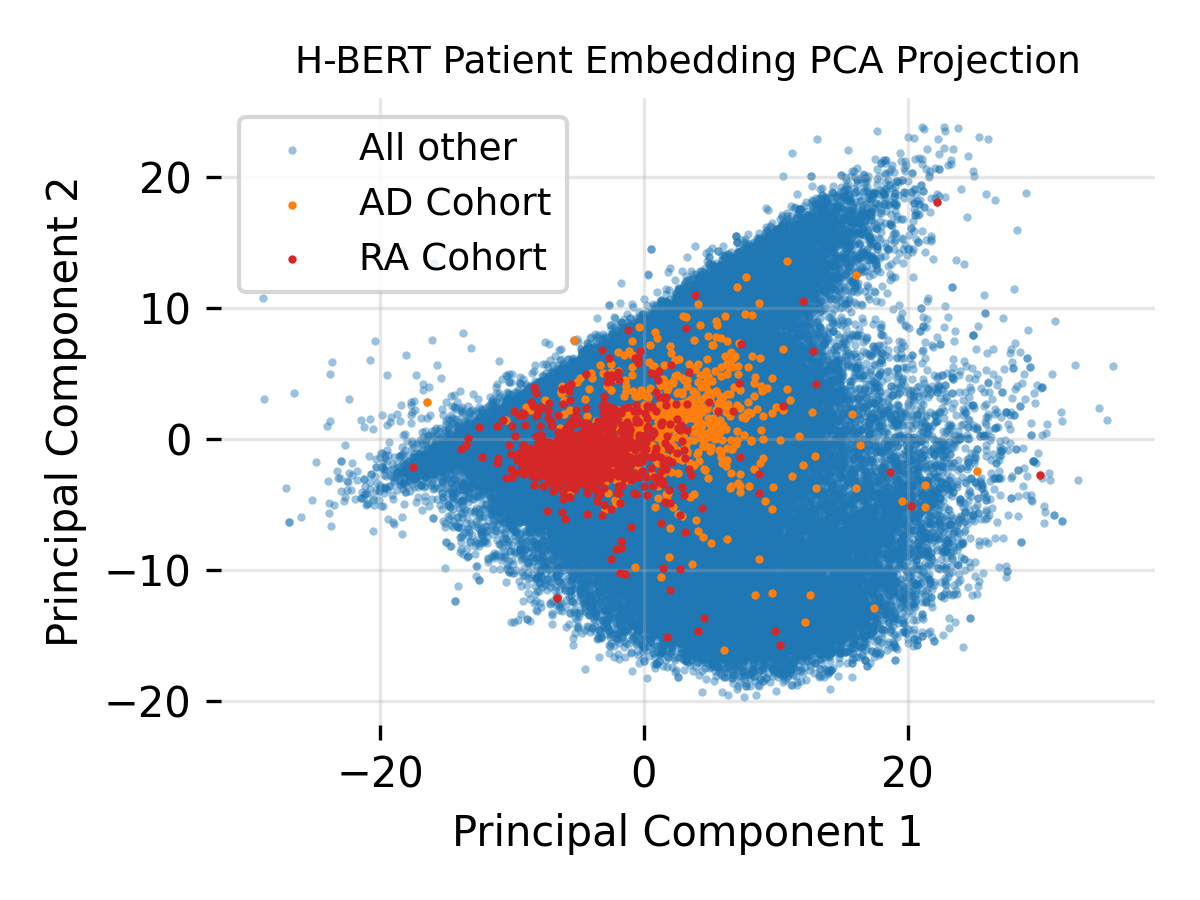}}
\end{figure}

\begin{figure}[htbp]
\floatconts
  {fig:pte_berto}
  {\caption{PCA projection of top two components of BERTO Patient Embeddings.}}
  {\includegraphics[width=1\linewidth]{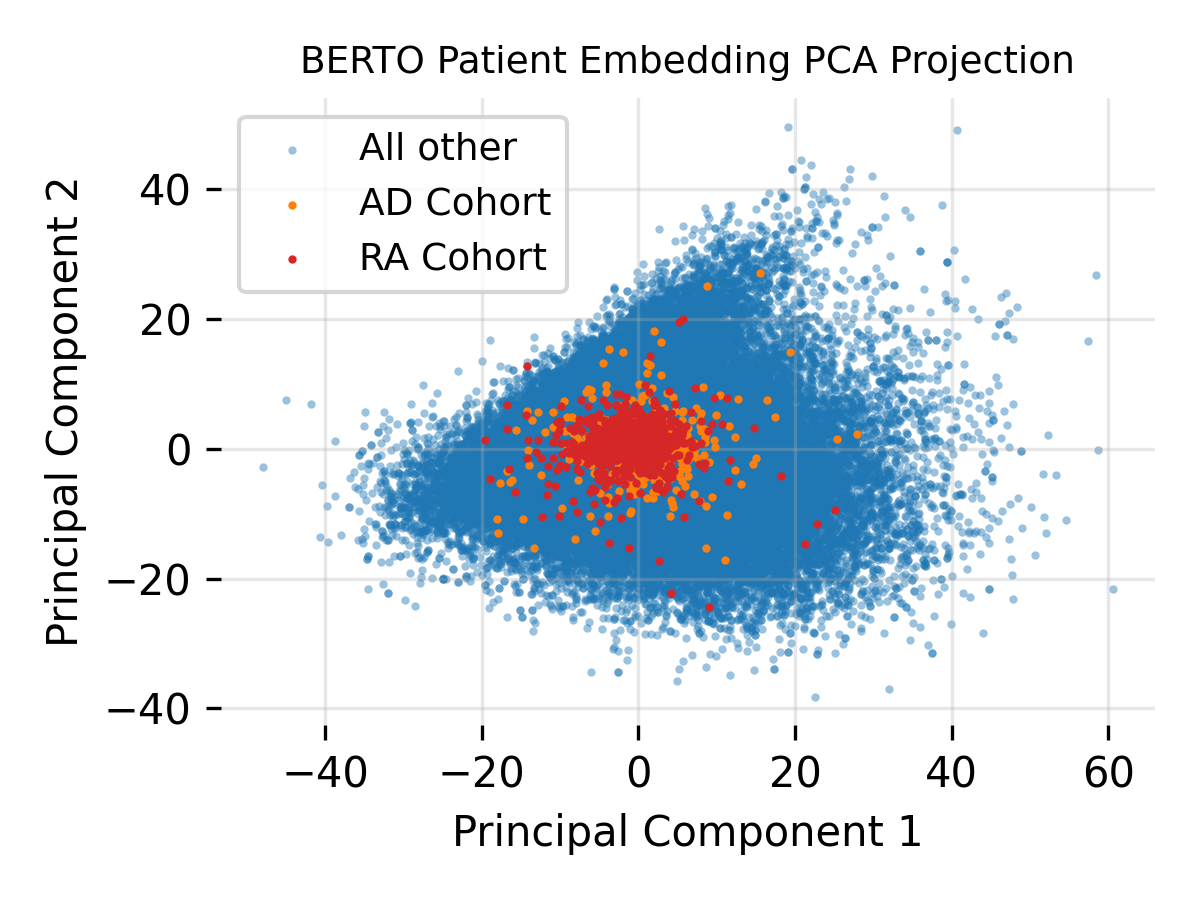}}
\end{figure}

\section{Conclusion}
Semantic decomposition alone yielded a benefit in terms of mean AUC and APS on 522 fine-tuning tasks as suggested by BERTO's improved performance compared to LEAFO.  H-BERT took that further by deploying a network architecture designed to exploit graphs.  Therefore, when inputs are symbols, combining semantic decomposition {and} structure appropriate networks (such as GAT) significantly improves learning.  In fact, {only this combined approach achieved meaningful separation of patient cohorts} as seen by projecting patient embeddings in PCA space.

This task provides some insight into the difficulty of predictive modeling from EHR.  Models capable of distinguishing heart disease and breast cancer are unable to separate two relatively common diseases that share somewhat similar etiology.  When distinct autoimmune diseases cannot easily be distinguished by a model, it is clear that actionable medical information is not being captured.

Hierarchical semantic decomposition also allows one to ask how much detail is needed for a given task by adjusting the maximum hierarchical token depth.  Perhaps suprisingly, visit-level embeddings can be aggregated to produce meaningful results without additional neural networks.  We do however envision a hierarchical BERT model where a second BERT is used to aggregate visit-level embeddings into a patient-level embedding, hoping BERT's dynamic sense of context will enable better results.

We intend to deploy this workflow on much larger insurance payer claims datasets, to enhance learning by using an order of magnitude more data.  By understanding the effect of data volume, we will be better able to diagnose the need for further architectural improvements.  We may also assess what can be learned given the information content of large medical datasets.  We also intend to expand this workflow by adding procedure, lab order, and provider specialty as inputs, all of which may be found in closed payer claims datasets along with EHR.

Two additional tasks envisioned are to label a patient's visits only (1) before or (2) after the first visit that activates the label.  Labelling before the class-activating visit enables explicit and unambiguous prediction of phenotype risk, whereas labelling after the class-activating visit enables phenotyping directly, and both tasks enhance explainability.

At the finest level, future work should include separate and also simultaneous pre-training and CCSR ablation to determine if and to what extent  pre-training, CCSR classification, and both tasks together improve performance in terms of mean AUC/APS \emph{and} separation of patient embeddings by phenotype as visualized in PCA-space.  As mentioned in the introduction, we aim to execute semi-supervised learning of phenotypes---to catch relevant patients that rules-based approaches miss and otherwise execute data science characterization of patients with embeddings similar to phenotypes of interest.

In the broadest of terms, we aspire to eventually find the optimal synergy between model architecture, input data volume and type, task set, and also factors such as weight initialization and training task order---to enable clinically-relevant and healthcare-relevant predictions and data science.


\bibliography{eqb_refs}

\appendix

\section{Raw Dataset}\label{apd:raw_data}

Our raw dataset provided approx.\ 240M diagnoses and 133M medication orders across 118M encounters and 2.2M patients.  Codes not otherwise presented as ICD-9 or ICD-10 were mapped to one of those systems using a map provided by \emph{anonymized data source}.  We then mapped all codes to ICD-10 and ATC-4, yielding an overall vocabulary containing 43,650 diagnosis and 1,936 prescription codes.

\section{Hierarchical Code Truncation}\label{apd:codeTrunc}
As mentioned above, the ICD-10 and ATC-4 medical code systems are hierarchical, in that codes may be shorter or longer according to the level of detail needed.  All codes in a system share that system's ``root'' token.  We define ``maximum hierarchy depth'' as the maximum number of semantic tokens retained; the retained tokens are the ones on the more general side of the limit (which always includes the ``root'' token), and the removed tokens are the more detailed side.  Using \figureref{fig:sem_decomp} as a reference, a maximum hierarchy depth for ICD-10 code ``I21.02'' of 2 leaves only two tokens:  ``root,'' and ``I00-I99.''  A maximum hierarchy depth of 4 for the same code results in 4 tokens:  ``root,'' and ``I00-I99,'' ``I21,'' and ``I21.0.''

The maximum hierarchy depth for ICD-10 in this work was 3, and the maximum hierarchy depth for ATC-4 in this work was 4.  This reduced the total number of diagnosis and prescription leaf tokens to 1,901 and 197, respectively.  The total number of hierarchical semantic tokens was 2,225, with 1,926 for diagnoses and 299 for prescriptions.

\section{Cohort Difinitions}\label{apd:cohorts}

``Cohorts'' of patients were constructed using conditional logic applied to EHR that define inclusion or exclusion from the cohort.  For example, if a patient has a diagnosis from a set of inclusive diagnoses twice, then that may be sufficient to include the patient.  Another example might be if a patient had two inclusive diagnoses, and inclusive medication, but not an exclusive medication.  The rules for our cohorts were developed iteratively with physicians knowledgeable of relevant clinical practice, which both improves our interpretation of EHR and confidence that the cohort members do have the underlying condition.  There is a tradeoff between the size of a resultant cohort and the confidence that cohort members have the underlying condition, based on how stringent the rules are.  Our rules tended to favor smaller cohorts with higher confidence.

\section{Hyperparameters}\label{apd:hyper}
All models used the same token embedding length of 64.  All models except LEAFO6 used 4 transformer blocks.  Transformer blocks each had 8 attention heads, 128 hidden nodes, transformer hidden node dropout probability of 0.4, and transformer attention head dropout probability of 0.1.  The GAT also used 8 attention heads, and averaged (rather than concatenated) the result.

Learning rate was $10^{-5}$ for all models and tasks, and training duration was 20 epochs.

\section{LEAFO Patient Embedding Visualization}\label{apd:leafo_pte}
\figureref{fig:pte_leafo} shows LEAFO does not produce patient-level embeddings for AD and RA patients that are demonstrably different.

\begin{figure}[htbp]
\floatconts
  {fig:pte_leafo}
  {\caption{PCA projection of top two components of LEAFO Patient Embeddings.}}
  {\includegraphics[width=1\linewidth]{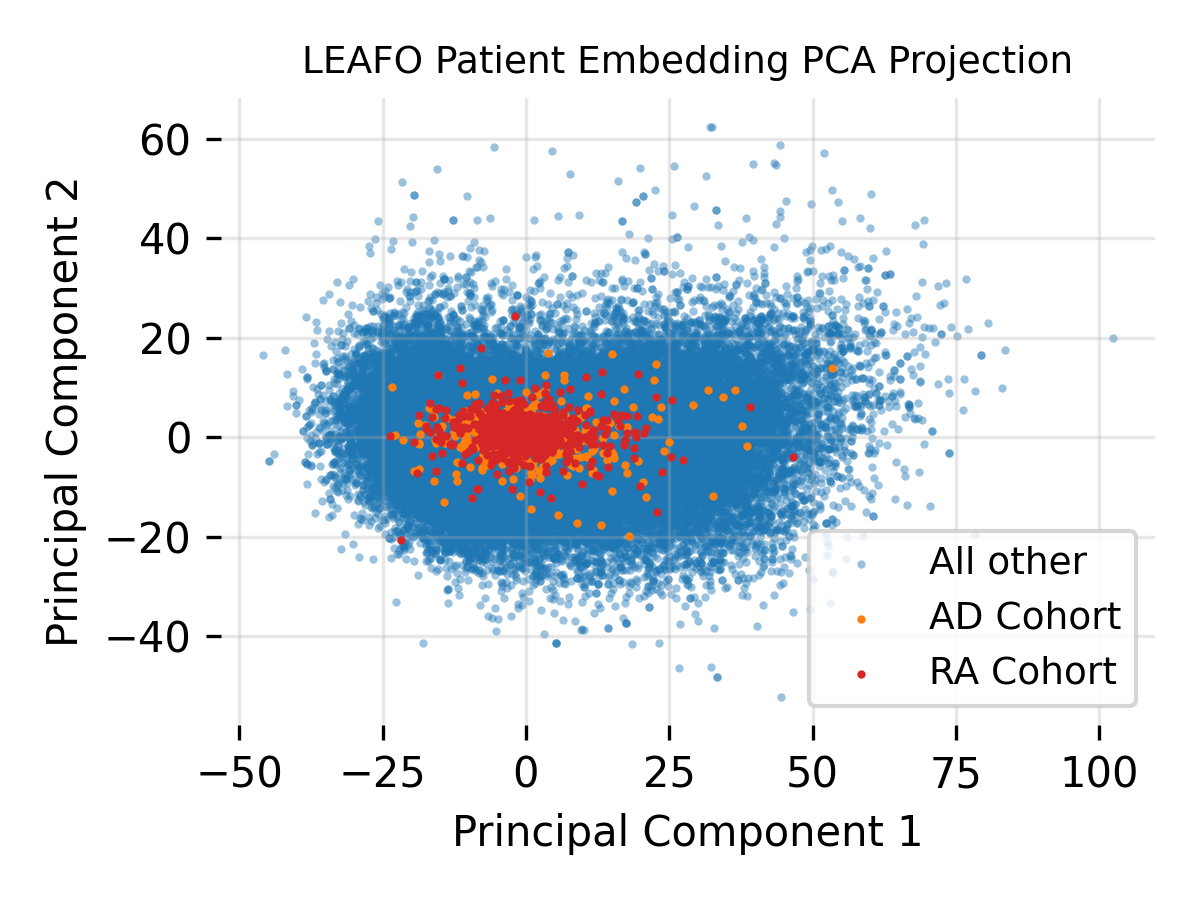}}
\end{figure}

\end{document}